\documentclass{article}

\usepackage{iclr2021/iclr2021_conference}
\iclrfinalcopy

\usepackage[numbers]{natbib}

\usepackage{algorithm}
\usepackage{algpseudocode}

\usepackage[utf8]{inputenc} 
\usepackage[T1]{fontenc}    
\usepackage{hyperref}       
\usepackage{url}            
\usepackage{booktabs}       
\usepackage{amsfonts}       
\usepackage{nicefrac}       
\usepackage{microtype}      
\usepackage{xcolor}         
\usepackage{mathtools}
\usepackage{enumerate}
\usepackage{multirow, multicol}
\usepackage{tabularx}
\usepackage{colortbl}
\usepackage{xcolor}
\usepackage{tabularray}

\usepackage[export]{adjustbox}

\usepackage{caption}
\usepackage{amsthm}

\usepackage{subcaption}
\usepackage{graphicx}

\usepackage{natbib}

\usepackage{amsfonts,amsmath,amssymb}
\usepackage{bbm}
\usepackage{xspace}

\usepackage{multicol}
\usepackage{enumitem}

\providecommand{\mc}[1]{\mathcal{#1}}

\usepackage{hyperref}
\usepackage{url}
\usepackage[hang,flushmargin]{footmisc} 

\title{A Statistical Turing test \\for generative models}

\author{Hayden Helm \\
Helivan Research \\
\texttt{hayden@helivan.io}
\And
Carey~E.~Priebe \\
Johns Hopkins University \\
\texttt{cep@jhu.edu}
\And
Weiwei Yang\\
Microsoft Research \\
\texttt{weiwya@microsft.com}
}

\usepackage{natbib}

\begin{document}

\begin{center}
\maketitle
\end{center}

\begin{abstract}
    The emergence of human-like abilities of AI systems for content generation in domains such as text, audio, and vision has prompted the development of classifiers to determine whether content originated from a human or a machine. 
    Implicit in these efforts is an assumption that the generation properties of a human are different from that of the machine. 
    In this work, we provide a framework in the language of statistical pattern recognition that quantifies the difference between the distributions of human and machine-generated content conditioned on an evaluation context. 
    We describe current methods in the context of the framework and demonstrate how to use the framework to evaluate the progression of generative models towards human-like capabilities, among many axes of analysis. 
\end{abstract}

\section{Introduction}

The development of foundation models whose representations and outputs can be used for various downstream tasks has facilitated unprecedented advancements in content generation across domains such as natural language \citep{GPT4-AOAI, Claude, PaLM2}, programming \citep{chen2021evaluating}, image \citep{clip}, and audio \citep{radford2022robust}, with some authors claiming glimpses of artificial general intelligence \citep{bubeck2023sparks}. 
These advancements, however, come with significant societal implications. 
For example, they present challenges to the employment prospects of knowledge workers \citep{eloundou2023gpts}, prompt a re-evaluation of the constructs of creativity and originality \citep{stable-diffusion-litigation}, and invite discourse on issues of privacy \citep{nytimesChatGPTBanned} and safety \citep{gpt4systemcard}.
Consequently, humanity is now tasked with understanding and navigating these technologies. 

As AI practitioners whose initiatives spurred the recent advancements in capabilities, we bear a special duty in this endeavor. 
Indeed, we believe that the development of evaluation frameworks that are robust enough to capture the evolution of applications within current domains and flexible enough to be extended to capabilities in new domains is imperative. 

Recent efforts such as the living benchmarks of holistic evaluation \citep{helm_not_hayden} for natural language processing tasks and the use of data kernels to understand differences in the embedding spaces of models \citep{duderstadt2023comparing} are important strides for understanding absolute and relative model capabilities, respectively. 
These efforts are in contrast with for-profit institutions releasing enthusiasm-laden reviews of their own work -- effectively claiming human-level capabilities across a suite of evaluation frameworks -- without including details necessary to reproduce their results and hence generating serious documentation debt for the community.

Beyond evaluating absolute or relative machine capabilities on specific downstream inference tasks, technologists have historically been interested in comparing machines to humans \citep{powers1998total}, with some authors arguing that the entire field of AI working towards the goal of emulating human-level ``intelligence" \citep{chollet2019measure}. 
The Turing test or imitation game, for example, is an evaluation framework used to determine whether a conversational agent is a human or a machine \citep{10.1093/mind/LIX.236.433}. 
In the originally proposed test, a person converses with the agent and determines if the agent is a human or machine once the conversation ends. 
If the agent is determined to be human, then it ``passed" the Turing test. Similar tests have been proposed in the vision domain \citep{doi:10.1073/pnas.1422953112}.
However, due to the of variability among human evaluators and the broad spectrum of tasks used for evaluation, there is a significant degree of ambiguity surrounding the criteria for passing. 

To address this ambiguity, we propose a statistical framework that quantifies the ability of a system to distinguish between human-generated and machine-generated content. 
We think of our framework as a set of statistical Turing tests for generative models.

\textbf {Human Detection Problem Statement:}
Let $ (x_{1}, y_{1}), \hdots, (x_{n}, y_{n}) $ be (content, label) pairs, where the label $ y_{i} \in \{0, 1\} $ determines whether the corresponding content $ x_{i} \in \mc{X} $ was generated by a machine (label 0) or a human (label 1). 
In the human detection problem, the goal of a classifier $ h: \mc{X} \to \{0, 1\} $ is to correctly determine whether content was generated by a human or a machine.

\textbf{Contributions:}
Our primary contribution is a statistical framework for human detection that can be used to quantify the detectability of a machine for a given \textit{human detection context} that is agnostic to content modality and generation task. 
The framework provides the language required to analyze important aspects of the human detection problem such as the progression of a family of models towards human-like abilities and the effectiveness of different classifiers for human detection. 
Secondary to the proposed framework is the contribution of a human detection method, \texttt{ProxiHuman}, that utilizes the geometry of a machine's embedding space to determine the origin (human or machine) of content. 




\section{Related Work}

Technologists and researchers have attempted to quantify a system's 
performance or ability -- in both absolute (e.g., classification accuracy) and relative (e.g., comparing a system's accuracy to human-level accuracy) terms -- countless times throughout our history \citep{hernandez2017measure}.
Quantifying absolute ability often relies on static datasets or evaluations crafted to test a system's ability to generalize and apply concepts previously learned to new examples or domains, such as HELM \citep{helm_not_hayden} for foundation models and the intelligence quotient for human psychometrics. Our work does not attempt to quantify absolute ability and is, instead, focused on relative ability by comparing the generative distributions of systems to the distribution of human content.

Quantifying relative ability relies on an existing baseline as reference. For example, in the original Turing test \citep{10.1093/mind/LIX.236.433} and its vision counterpart \citep{doi:10.1073/pnas.1422953112} a human judge uses past experience conversing with other humans as a basis for their final judgement. 
There is considerable variance in a randomly selected human judge's perceived previous experience and these tests, largely, have fallen out of favor. 
For example, the Loebner prize \citep{powers1998total} was terminated in 2020 after more than 20 years of offering a never-awarded grand prize of \$100,000 to the first team to produce a machine that could convince more than half of the judges that it was human.

Recently, Zhang et al. directly compared a model's ability to distinguish between human generated content and machine generated content relative to human judges \citep{zhang2022human} and found that both human judges and machine judges fail to classify data from particular domains sufficiently. The authors refer to their evaluations as a set of Turing tests.
One of the purposes of the current work is to translate the goals of competitions such as the Loebner Prize and of evaluations such as in Zhang et al. into a statistical framework with quantitative measures of distinguishability. We foresee a ``statistical-Loebner Prize" where a grand prize is awarded to a machine that is $ \tau $-undetectable (defined in Section \ref{sec:framework}) for a sufficiently large sample space and for $ \tau $ sufficiently small.



Intertwined with efforts to quantify absolute and relative intelligence is the generation and detection of adversarial ``deepfakes" \citep{nguyen2022deep}, which can be exploited to disseminate misinformation, propagate fake news, or execute scams. The deepfake literature has mainly focused on the video and image domains as these were the only modalities where someone could use off-the-shelf architectures and training methods to produce convincingly fake content. See surveys \citep{Verdoliva_2020, nguyen2022deep} for detailed descriptions of state-of-the-art generation and detection methods in vision. Recently, however, improved capabilities and availability of models, such as through HuggingFace's Model Hub \citep{wolf2019huggingface}, has resulted in an increased interest in deepfake detection and creation in the audio \citep{khanjani2023audio} and natural language \citep{mitchell2023detectgpt} domains. 

With the increased popularity and accessibility of the generative models, there is increased activity in ``abusing" them -- for students to use it as a shortcut in their academic endeavors \citep{nytimesUniversityAIBan} or for lawyers to use it to replace their case diligence processes \citep{lawyers-chat-gpt}. To combat ``abuses" from users of its ChatGPT, OpenAI released its own GPT detector \citep{AITextClassifier} that has since been removed from their services due to high misclassification rates. Indeed,  detectors  typically use information from the model used to generate the content to perform the detection \citep{ippolito-etal-2020-automatic, AITextClassifier} and experience performance degradation when the generative model is not known \textit{a priori} and available or the content is otherwise out of domain \citep{bakhtin2019real, uchendu-etal-2020-authorship, mitchell2023detectgpt}.

Our proposed framework endows a context with a single scalar for each model. The scalar represents the futility of human detection and unifies deepfake generation and detection efforts across multiple modalities. 
Importantly, the framework provides the necessary language to compare models within and across model families, to compare multi-modal models to uni-modal models, to compare the same detection algorithm instantiated with different hyperparameters, to compare different detection algorithms on different datasets, and to compare the same detection algorithm across datasets, among others.

\section{Human detection in the language of statistical pattern recognition}
\label{sec:framework}

The human detection problem is an instance of a more general statistical pattern recognition problem~\citep[Chapter~1]{devroye2013probabilistic}: Given labeled data $ \{(X_{i}, Y_{i})\}_{i=1}^{n} \in \left(\mathcal{X} \times \{0,1\}\right)^{n} $ assumed to be $i.i.d.$ samples from a classification distribution $ \mathcal{P} $, construct a function $ h $ that takes as input an element of the sample space $ \mc{X} $ and outputs an element of $ \{0, 1\} $. Herein we assign the label ``0" to a machine and the label ``1" to a human. The classifier $ h $ is evaluated with respect to a loss function $ \ell: \{0, 1\} \times \{0,1\} \to \mathbb{R}_{+} $ that compares the predicted label $ h(X) $ and the true label $ Y $. We define the risk of $ h $ as the expected loss of $ h $ with respect to $ \mc{P} $, or $ R(h; \mc{P}, \ell) = \mathbb{E}_{\mc{P}}\left[\ell(h(X), Y)\right] $. When possible, we abbreviate $ R(h; \mc{P}, \ell) $ as $ R(h) $ for convenience. The classifier $ h $ is preferred over $ h' $ if $ R(h) < R(h') $. We let $ \mc{H} = \{h: \mc{X} \to \{0, 1\}\} $ denote a set of classifiers and define $ h^{*} $ as the classifier that minimizes the risk within $ \mc{H} $ and $ h^{**} $ as the classifier that minimizes the risk amongst all functions from $ \mc{X} $to $ \{0, 1\} $.

The classification distribution $ \mc{P}=\pi f_{0} + (1-\pi) f_{1} $ can be decomposed into the weighted sum of the machine's class conditional density $ f_{0} $ and the human's class conditional density $ f_{1} $. 
Given the class 0 prior $ \pi $, we define the \textit{chance-level} classifier $ h^{c} $ as the classifier that is independent of the input $ x \in \mc{X} $ and returns $ 0 $ with probability $ \pi $ and $ 1 $ with probability $ 1 - \pi $. 
For there to exist a classifier that is
useful for human detection, it is necessary that the class conditional distributions $ f_{0} $ and $ f_{1} $ be sufficiently distinct. 

Classifiers used in the context of human detection often operate on transformations of the original input space \cite{gehrmann-etal-2019-gltr, mitchell2023detectgpt}. We let $ t: \mc{X} \to \mc{X}' $ denote a transformation of the original content space and note the inequality
\begin{align}
\label{information-processing}
    R(h^{c}; \mc{P}, \ell) \ge R(h_{t}^{**}; \mc{P}, \ell) \ge R(h^{**}; \mc{P}, \ell),
\end{align} where $ h_{t}^{**} $ is the optimal classifier after transforming the original inputs $ x \in \mc{X} $ with $ t $. The chain of inequalities described in Eq. \eqref{information-processing} is a restatement of the information processing lemma \citep{beaudry2012intuitive}. Importantly, Eq. \eqref{information-processing} does not give any hint of the magnitude of either $ R(h_{t}^{**}; \mc{P}, \ell) - R(h^{**}; \mc{P}, \ell) $ or $ R(h^{c}; \mc{P}, \ell) - R(h_{t}^{**}; \mc{P}, \ell) $ without any explicit assumptions on the relationship between $ \mc{P}, \ell, t $ and $ \mc{H} $.

For facilitating analysis of the human detection problem along relevant axes, we define the \textit{human detection context} $ \mc{C} $ as a sextuple consisting of a sample space $ \mc{X} $, human-generation class conditional distribution $ f_{1} $, class mixing coefficient $ \pi $, loss function $ \ell $, transformation $ t $, and a set of classifiers $ \mc{H} = \{h: t(\mc{X}) \to \{0, 1\}\} $. We say that a machine, parameterized by its content generation distribution $ f_{0} $, is $ \tau $-undetectable for context $ \mc{C} = (\mc{X}, f_{1}, \pi, \ell, t, \mc{H}) $ if 
\begin{align}
\label{eq:tau-detectable}
    1 - \frac{R(h_{t}^{*}; \mathcal{P}, \ell)}{R(h^{c}; \mathcal{P}, \ell)} \le \tau.
\end{align} Colloquially, for a machine to be $ \tau $-undetectable for context $ \mc{C} $ for small $ \tau $ means that the human-generation distribution and machine-generation distribution are effectively indistinguishable and that the machine has ``passed" the statistical Turing test for $ \mc{C} $. 

The normalization by the performance of the chance-level classifier in Eq.\ \eqref{eq:tau-detectable} allows for (na\"ive) comparison of contexts. 
For example, if there exists a model $ f_{0} $ for context $ \mc{C} $ that is $ \tau$-undetectable, but there does not exist a model for context $ \mc{C}' $ that is $ \tau$-undetectable, then we may reason that $ \mc{C} $ is a harder context than $ \mc{C}' $.


\subsection{The human detection context}

We now highlight how changing elements in the human detection context can change the scope of a detectability claim and other important properties of the proposed framework.

\textbf{The sample space $ \mc{X}$ and human class conditional distribution $ f_{1} $.} The sample space $ \mc{X} $ is the set of the content under consideration and can be as general as ``the set of all content ever" or as specific as ``science-fiction writing in the style of Isaac Asimov". 
The sample spaces populated by content from recent undetectability claims are less general than ``the set of all written content" but general enough to capture our imagination. 

The human class conditional distribution $ f_{1} $ assigns mass or measure to elements or regions of the sample space. The exact mapping depends on the type of detection task and there are potentially multiple ``human" distributions for a given sample space. For example, the ``human" distribution can represent the distribution of content from an individual (e.g., an image derived from a painting by Claude Monet), a group of individuals (e.g., an image derived from a painting by an impressionist), or a standardized content producing procedure (e.g., from a class teaching students to paint like the impressionists). 
Holding all other elements of the context constant, a machine can be $ \tau $-undetectable for one type of human distribution and not for the others. It is thus important to attempt to characterize the human distribution for which we claim a machine is $ \tau $-undetectable.

For a distribution $ f_{1} $ on the sample space $ \mc{X} $, we say the distribution $ f_{1}' $ is an extension of $ f_{1} $ to $ \mc{X}' $ if $ X | x \in \mc{X} \sim f_{1} $ for $ X \sim f_{1}' $. 
Notably, if a machine is $ \tau $-undetectable for context $(\mc{X}, f_{1},\pi,\ell,t,\mc{H}) $ and $ \tau'$ -undetectable for context $ (\mc{X}',f_{1}',\pi',\ell',t',\mc{H}') $ and $ \mc{X} \cap \mc{X} = \varnothing $ then there exists a context  $ (\mc{X} \cup \mc{X}',f_{1}'',\pi'',\ell'',t'',\mc{H}'') $ such that the machine is max($ \tau $, $\tau'$)-undetectable. 
The constructive proof of this claim relies on letting $ f_{1}'' $ be an extension of $ f_{1} $ to $ \mc{X}' $ and an extension of $ f_{1}'' $ to $ \mc{X} $. 

The modularity of $ \tau $-undetectability for contexts with non-intersecting sample spaces allows researchers to make claims about the relative abilities of models at various specificities. For example, a highly specialized science fiction-writing model may be more human-like for writing about humanity's migration to nearby solar systems than a general purpose foundation model. The highly specialized model would have a smaller $ \tau $-undetectability in this context. Including other sample spaces in the evaluation, however, would likely cause the difference in $ \tau $-undetectability of the models to change signs. In this sense, the framework enables comparing both highly specific generative ability and general generative ability.

\textbf{The class conditional prior $ \pi $ and loss function $ \ell $.} The class conditional prior $ \pi $ is proportional to the prominence of machine generated content. In situations such as writing and art competitions, we'd hope that $ \pi $ is much smaller than in situations such as spam email generation. In reality, however, the cost of a machine to produce content is much cheaper than the cost of a human to produce content and so it may be sensible to assume that $ \pi \approx 1 $ for most situations where the content generation process is not visible.

The loss function $ \ell $ determines how ``costly" misclassifying a piece of human generated content as originating from a machine and vice versa. Different use cases of human detectors will require changing the loss function appropriately. For example, service providers that value their user's experience will not want to risk upsetting a user by labeling them as a machine and requiring them to complete more captchas. On the other hand, government agencies that want to ensure that the user is an actual person so as to avoid fraudulent activity may want to weigh the misclassification of machine generated content highly. 

There is a duality in the relationship of the class conditional prior $ \pi $ and loss functions $ \ell $ that assign positive weights $ \alpha $ (for misclassifications of human generated content) and $ \beta $ (for  misclassifications of machine generated content) such that $ \alpha + \beta = 1 $. That is, for a given sample space $ \mc{X} $, class conditional distributions $ f_{0} $ and $ f_{1} $, classifier $ h $ and fixed risk $ R $ there are multiple $ (\pi, \alpha, \beta) $ triples such that $ R(h; \mc{P}, \ell) = R $ \citep{bickel2015mathematical}. As with the restatement of the information processing lemma above, it is not possible to comment on the relationship between the detectability of a single machine within different contexts without explicit assumptions on the relationship between $ \mc{P}, \ell, t $ and $ \mc{H} $.

\textbf{The transformation $ t $ and set of classifiers $ \mc{H} $.} The transformation $ t $ maps an input modality such as an image, segment of natural language, multi-channel time series, or graph to a space more suitable to analysis such as a Euclidean space $ \mathbb{R}^{d} $. 

The elements in the set of classifiers $ \mc{H} $ map transformed inputs to $ \{0, 1\} $. The relationship between $ t $ and $ \mc{H} $ can be quite complicated. For simplicity we assume that a transformation maps the original input to a scalar and that the set of classifiers is parameterized by thresholds $ c $ such that if $ t(X) < c $ then $ h_{c}(t(X)) = 0 $ and $ h_{c}(t(X)) = 1 $ otherwise. This constraint on the relationship between $ t $ and $ 
\mc{H} $ includes popular classifiers such as decision tress and forests (via thresholding the estimated posterior), support vector machines (via thresholding the distance to support vectors), and artificial neural networks (via thresholding the softmax).




\section{Evaluating empirical $ \tau $-undetectability}

In this section we demonstrate how to use the human detection statistical framework to make detectability claims. To this end, we first describe five different human detection methods as transformations of the original input space to a single scalar. We then describe four natural language datasets and characterize their human-conditional distribution. Finally, we measure the detectability of OpenAI's ChatGPT (``GPT3") and GPT4 for producing content related to the four datasets.


\subsection{Example Detection Methods}

We describe the transformation $ t $ for four different human detection methods: \textit{Likelihood}, \textit{LogRank}, \textit{DetectGPT}, and \textit{ProxiHuman}. The set of classifiers $ \mc{H} $ associated with each transformation is the set of classifiers parameterized by a single scalar threshold.  

The four transformations we evaluate were all initially developed and analyzed under the ``white-box" setting \citep{gehrmann-etal-2019-gltr} where we have access to non-output functions from the model. We let $ \theta $ parameterize the non-output functions. All methods require a function $ p_{\theta} $ that evaluates the likelihood of an observed sequence of tokens except for ProxiHuman. ProxiHuman requires an embedding function $ g_{\theta} $ that maps a sequence of tokens to a vector space. 

\textbf{Likelihood} \citep{solaiman2019release} is a simple baseline based on the assumption that a generative model will more-often-than-not choose tokens that have high likelihood from the perspective of $ p_{\theta} $. Machine generated text will have a higher likelihood (on average) than human generated text.
For a given piece of content $ x $ split into tokens $ x = (x^{(1)}, \hdots, x^{(n)}) $, the transformation $ t $ is for Likelihood is thus $ \frac{1}{n}\sum_{i=1}^{n} \log p_{\theta}(x^{(i)}) $.


\textbf{LogRank} \citep{ippolito-etal-2020-automatic} also assumes that machine generated text will contain tokens that have a high likelihood from the perspective of $ p_{\theta} $. Instead of simply taking the average likelihood of the sequnce of tokens, LogRank evaluates $ p_{\theta} $ for every token in the vocabulary in the provided context and records where the observed token falls in the list. The transformation for LogRank is thus $ \frac{1}{n} \sum rank_{i}(p_{\theta}(x^{(i)}) $ where $ rank_{i} $ finds the rank of the likelihood of $ x^{(i)} $ relative to the rest of the vocabulary.

\textbf{DetectGPT} \citep{mitchell2023detectgpt} is a zero shot human detection method
and is based on the observation that ``samples from a source model typically lie in areas of negative curvature" of 
$ \log p_{\theta} $. In contrast, text written by a human is less likely to exist in these regions. 

To leverage this observation into a human detection method, the authors propose perturbing the observed content $ x $ locally by masking tokens of $ x $ and filling the masked tokens with a potentially-different model to produce a new piece of content $ \tilde{x} $. Given the different assumptions on the curvature of $ \log p_{\theta} $ for humans and machines, the absolute difference between the original $ \log p_{\theta}(x) $ and the average of $ \log p_{\theta}(\tilde{x}) $ should be small for machine generated content and large for human generated content. The transformation $ t $ is thus $ p_{\theta}(x) - \mathbb{E}_{\tilde{p}}\left[\log p_{\theta}(\tilde{x})\right] $ with $ \tilde{x} \sim \tilde{p} $ and where $ \tilde{p} $ is dependent on the original content $ x $ and the proportion of original content that is masked.

\textbf{ProxiHuman} is a novel zero shot human detection method that is heavily inspired by DetectGPT. We assume that the user has access to an embedding function $ g $ that maps content $ x $ to an element in $ \mathbb{R}^{d} $ and that the embeddings from human-generated content are ``far" from the region of space in which the embeddings from machine-generated content exist. We further assume that the user has access to a dissimilarity measure $ d $ suitable for the embedding space. 

As with DetectGPT, we translate that assumption into a method by perturbing the original content $ x $ via mask filling. 
Unlike DetectGPT, our assumption leads naturally to considering the distance between $ g_{\theta}(x) $ and the manifold on which the sampled $ g_{\theta}(\tilde{x}) $ exist. 
We can approximate this distance using the average distance between $ g_{\theta}(x) $ and its $ k $-nearest neighbors amongst the perturbed samples. 
The transformation $ t $ for ProxiHuman is thus $ \frac{1}{k} \sum_{i=1}^{N} d(g(x), g_{\theta}(\tilde{x}_{i})) \cdot \mathbbm{1}\{g_{\theta}(\tilde{x}_{i}) \in \mathcal{N}_{k}(g_{\theta}(x), d)\} $ where $ \mathcal{N}_{k}(g_{\theta}(x), d) $ is the set of $ g_{\theta}(x) $'s  $ k $ nearest neighbors with respect to $ d $ and $ \mathbbm{1} $ is the indicator function. 
See Algorithm \ref{alg:Proxihuman} in the appendix for a detailed algorithmic description.

Since the aim of human detection is to determine the origin of the content, our evaluations are in the ``black-box" setting where we do not have access to functions to evaluate the non-output functions from the model. In the black-box setting, we replace the model that generated the content with a surrogate model to generate perturbed content and to evaluate the likelihood of observing the original and perturbed content (for DetectGPT) or to evaluate the embeddings of the original and perturbed content (for ProxiHuman). In the experiments below we use the T5-large model \citep{raffel2020exploring} as the surrogate perturbation model and mask a random 10\% of the original tokens for each perturbation. We use GPT2 \citep{radford2019language} as the surrogate likelihood evaluation model for Likelihood, LogRank, and DetectGPT and MPnet \citep{song2020mpnet} as the surrogate embedding model for ProxiHuman evaluations. 




\subsection{Dataset and prompt descriptions}

\label{subsubsec:nlp-datasets}
We consider four different natural language datasets: GPT-wiki-intro (Wiki-Intro)\citep{aaditya_bhat_2023}, XSum \citep{narayan-etal-2018-dont},  WritingPrompts \citep{fan-etal-2018-hierarchical}, and PubMed QA \citep{jin-etal-2019-pubmedqa}. Each dataset contains examples of human generated content that we use as elements generated from the human class conditional distribution $ f_{1} $. Descriptions of the data, the human class conditional distribution $ f_{1} $, and the method we use to sample elements from the machine class conditional distribution $ f_{0} $ for the four datasets are as follows:

\textbf{Wiki-Intro} on the Huggingface Hub is a collection of Wikipedia introduction paragraphs and a corresponding text string that can be used to prompt a model to generate a Wikipedia-style introduction \citep{aaditya_bhat_2023}. For a given topic, the text string includes instructions to generate a 200 word introduction and includes the first seven words of the actual introduction as a warm start. To approximate the human conditional distribution, we randomly sampled 100 of the provided real Wikipedia introductions. In this case, the human conditional distribution is the distribution of highly edited content with (potentially) multiple contributors. To approximate a machine's conditional distribution, we randomly sampled another 100 of topics and prompted a machine with the provided text string for the selected topics. 

\textbf{XSum} is a collection of articles from  British Broadcasting Company and their corresponding single sentence summary \citep{narayan-etal-2018-dont}. The summary is typically written by the author of the articles and the articles cover a wide variety of topics. To approximate a machine's conditional distribution, we randomly sampled another 100 stories and prompted the machine to finish the news story given the first 30 tokens and the requirement that the response is a certain length. Stories that contained gory or otherwise explicit details were omitted.

\textbf{WritingPrompts} is a collection of creative writing prompts and corresponding responses provided by the Reddit community \citep{fan-etal-2018-hierarchical}. To approximate the human conditional distribution, we randomly sampled 100 of the provided responses. The content is highly individualized and the studied human conditional distribution is a mixture of the distributions of multiple individuals. To approximate a machine's conditional distribution, we randomly sampled another 100 responses and prompted the machine to finish the story given the first 30 tokens and the requirement that the generated story is a certain length.

\textbf{PubMed QA} is a collection of medical questions derived from PubMed abstracts \citep{jin-etal-2019-pubmedqa}. The authors construct a question (e.g., ``Does a factor influence the output?") based on the title and abstract and generate a corresponding long form answer (e.g. ``The study seems to indicate that the factor influences the output.") and short form answer (e.g., ``yes"). We randomly sampled 100 questions and used the long form answers to approximate the human conditional distribution. For the machine conditional distribution, we randomly sampled 100 questions, prompted the machine with the question, and required the machine to return an answer with the same number of tokens as the true-but-unseen human long form answer.

The actual prompts used to produce the machine-generated content for each dataset are shown in Table \ref{tab:prompts}.

\begin{table}[t]
    \centering
    \begin{tabular}{|c|c|}
    \hline
    Dataset & \multicolumn{1}{c|}{Prompt design} \\
    \hline \hline
    \multirow{2}{5em}{Wiki Intro}     &  200 word wikipedia style introduction  \\
    & on `\{topic\}' \{start\_prompt\} \\
    \hline
    \multirow{2}{5em}{XSum} 
    & Please continue this news story \\ 
    & using \{target\_answer\_length\} words. \{start\_prompt\}: \\
    \hline
    \multirow{2}{5em}{Writing Prompts}     & You are a creative writer. Please write a compelling story \\
    & of around \{target\_answer\_length\} words based \{start\_prompt\} \\
    \hline
    \multirow{3}{5em}{PubMed QA}     &  You are an AI model trained to share general medical knowledge. \\
    & Please answer the medical question in \{target\_answer\_length\} words \\ 
    & and be as informative as possible.\\
    \hline
    \end{tabular}
    \caption{The prompt designs used to prompt GPT3 and GPT4 when generating machine-generated content for the four datasets described in Section \ref{subsubsec:nlp-datasets}}
    \label{tab:prompts}
\end{table}

\subsection{Evaluation and results}
\label{subsec:eval-and-results}
For all contexts we fix the mixing coefficient $ \pi $ to be 0.5 and the loss function $ \ell $ to be 0-1 loss. We do not have access to the true-but-unknown class conditional distributions $ f_{0} $ and $ f_{1} $ to properly measure the risk $ R(h) $ associated with each threshold. Instead, we report empirical $ \tau $-undetectability which replaces $ R(h) $ with the empirical risk $ \widehat{R}(h) $.\footnote{Since the transformations we study do not use supervised information to learn the transformation $ t $, the empirical $ \tau $-undetectability will be a reasonable estimate of the true $ \tau $-undetectability. If using supervised transformations prone to overfitting we recommend studying the cross validation-based analog of empirical $ \tau $-undetectability.} A higher empirical $ \tau $-undetectability means that the content from the model is more easily distinguished from human content. The empirical $ \tau $-undetectabilities of GPT3 and GPT4 are given in Table \ref{tab:nlp-table}.

\begin{table}[t]
\small
\centering
\begin{tabular}{|c||c|c|c|c|c|c|c|c||c|c|}
\hline
      & \multicolumn{8}{c||}{Empirical $ \tau $-undetectabilties} 
      & \multicolumn{2}{c|}{}
      \\
      \hline
      & \multicolumn{2}{c|}{PubMed} & \multicolumn{2}{c|}{W. Prompts} &
      \multicolumn{2}{c|}{XSum} & 
      \multicolumn{2}{c||}{Wiki} & \multicolumn{2}{c|}{Max}\\
     \hline\hline
     GPT & 3 & 4 & 3 & 4 & 3 & 4 & 3 & 4 & 3 & 4\\
     \hline
     Likelihood 
& \textbf{0.60} & 0.31 
& \textbf{0.95} & \textbf{0.63} 
& \textbf{0.88} & 0.64 
& \textbf{0.96} & \textbf{0.60} 
& 0.96 & 0.64 \\
    Log Rank
& 0.58 & 0.33 
& 0.94 & 0.54 
& 0.72 & \textbf{0.66} & 
\textbf{0.96} & 0.55 
& 0.96 & 0.66 \\
     Det. GPT 
& 0.49 & 0.21 
& 0.60 & 0.33 
& 0.42 & 0.24 
& 0.68 & 0.52
& 0.68 & 0.52\\
     Pr. Human 
& 0.41 & 0.26 
& 0.26 & 0.16 
& 0.31 & 0.12 
& 0.64 & 0.29 
& 0.64 & 0.29\\
     \hline \hline
     Average 
& 0.52 & 0.29 
& 0.72 & 0.45 
& 0.52 & 0.40 
& 0.75 & 0.45 
& 0.82 & 0.54 \\
     \hline
     Avg. gap 
& \multicolumn{2}{c|}{0.23} 
& \multicolumn{2}{c|}{0.27} 
& \multicolumn{2}{c|}{0.12} 
& \multicolumn{2}{c||}{0.30} & \multicolumn{2}{c|}{0.28} \\
     \hline
\end{tabular}
\caption{The empirical $ \tau $-undetectabilities for five different transformations in 4 different NLP-based human detection contexts for GPT3 and GPT4. 
The transformation that leads to \{GPT3, GPT4\} being most detectable for each context is \textbf{bolded}. 
The maximum $ \hat{\tau} $ for each transformation -- representing the detectability of the model for the union of the 4 modular contexts -- are given in the second column from the right and the rightmost column for GPT3 and GPT4, respectively.
}
    \label{tab:nlp-table}
\end{table}


\textbf{GPT4 produces more human-like content than GPT3 in all four contexts.} 
For each transformation the content produced by GPT4 is harder to distinguish from human content than the content produced by GPT3. Across all transformations, the average detectability gap is greater than 0.12 for every dataset we studied. Independent of improvements of absolute evaluation metrics such as HELM and improvements in the accessbility of models through public APIs, the more human-like content being produced by GPT4 has likely contributed to the recent widescale adoption of large language models.

\textbf{The gap between GPT3 and GPT4 is smallest for news story generation.} The gap in average detectability between GPT3 and GPT4 is as high as 0.3 and as low as 0.12. Interestingly, the gap is smallest when producing news stories. We think that the ``small" gap is due to the nature of news -- it, by definition, involves never-before-seen events of which a language model will have no knowledge of. The improvement between GPT3 and GPT4 is likely due to improvements in the model's use of English in general and not due to improvements in its knowledge base. This oberservation is in line with the extensive analysis of the effect of outdated pre-training data on model performance on benchmark datasets \citep{longpre2023pretrainers}.

\textbf{GPT\{3, 4\} are least detectable when producing medical content.} 
From the perspective of how quickly communicating information can change, medical Q\&A is perhaps on the other extreme from news stories. The conventional wisdoms of medical practice -- and, more importantly for our study, how they are communicated -- is relatively slow to change, and we think this contributes to the detectability of GPT3 and GPT4 being least detectable in this context.

The human distribution for the long-form answers of the PubMed Q\&A might be drastically different than the human condition for a particular medical professional and that our analysis gives no insights on the ability of GPT3 or GPT4 to produce content that is similar to any particular medical professional.


\textbf{GPT4 is still detectable (in general).} Despite the large detectability decrease from GPT3 to GPT4, we interpret the average maximum $ \hat{\tau} $ for GPT4 being 0.54 as the model's content still being distinguishable from human content at the population level. This, however, does not imply that a human detection method will correctly identify the origins of an \textit{individual} piece of content. In fact, with a average maximum detectability of 0.54, we discourage the deployment of human detectors for pieces of individual content.   




\textbf{Simple rules survive.} While each method we studied was developed in the white-box setting where the transformation has access to non-output functions of the generative model, there is a large difference between the performances of the five detection methods. In particular, the methods that rely on complicated geometric assumptions of either the likelihood function (DetectGPT) or embedding space (ProxiHuman) perform much worse than simpler assumptions on the same objects such as the Likelihood and Log Rank methods. We interpret the relative difference in detectability as related to the ``closeness" of the surrogate model to the original generative model. Surrogate models that are more similar to the generative model are more likely to retain the geometric properties exploited by methods like DetectGPT and ProxiHuman. Surrogate models that are less similar to the generative model will likely still have first or second order functions -- such as the likelihood of observing the token or the rank order of the likelihoods -- that are similar to the generative model but have dissimilar higher order functions. In this sense, ``simple rules survive" \citep[Chapter 5]{devroye2013probabilistic}.

\section{Limitations and future work}

There are limitations of our current work that motivate future work. For example, it is well known that the quality of the output from a generative model is dependent on the prompt \citep{liu2021pretrain}. It may be the case that with proper prompt engineering, the $ \hat{\tau} $ of the models we studied are actually lower than what we report. Indeed, for a given context the $ \hat{\tau} $ we report serves as an upper bound to the $ \hat{\tau} $ with optimal prompt engineering. 

Further, we did not fully explore how changing different elements of the human detection context -- such as $ \pi $ or $ \ell $ -- can affect detectability claims. We think this to be an important set of analysis as the majority of use cases for human detection methods will not have $ \pi = 0.5 $ or will not want to weigh Type I and Type II errors equally. 

Similarly, we did not include other model families such as PaLM, LLaMA, Claude, etc. in our experiments. Nor did we explore the effects of quantization or fine-tuning on detectability. Each of these directions are necessary to fully understand the landscape of the human-ness of the content that is being produced of foundation models.

Finally, we only experimentally investigate $ \hat{\tau} $-undetectability in the natural language domain. The framework should be applied to any domain where generative models are available. Of notable omission in our analysis are the image, conversational and audio domains. While we expect the same general trends (e.g., more recent models are less detectable than less recent models and there are some sample spaces that the models are less detectable than others) to hold in these domains, a proper empirical study is necessary.

\section{Discussion}
In this work we introduced a statistical framework for evaluating human detection methods for a given human detection context. The human detection context is a sextuple of objects that define the classification problem. Changing a single element of the human detection context can change the meaning and scope of a detectability claim. We view our framework as a step towards objectively quantifying the progress and human-likeness of generative models across diverse domains of content.

As an example of how to use the framework to study different aspects of the human detection problem, we evaluated five different human detection methods that utilize different assumptions for the difference between human and machine generated content. 
We showed GPT4 is more human-like than GPT3 for all four methods and four datasets, highlighted that GPT3 and GPT4 are relatively adept at answering medical questions in a human-like manner and that the gap in human-ness between GPT3 and GPT4 is smallest for news story generation. We also argued that GPT4 is detectable in a general sense but that it is likely unwise to deploy a human detector with the current landscape of models. We lastly commented on the fact that the detection methods that leverage simple functions of the surrogate model outperform detection methods that rely on more complicated relationships between human and machine generated content in the black-box setting.


Our human detection results are not intended to be surprising nor state-of-the-art. They are intended to be illustrative of the types of analysis that can be done under the proposed framework when controlling for different aspects of the human detection context. For example, we demonstrated that it is possible to compare models within the same context, to study the effect of different transformations on detectability, to comment on the relative difficulty of generation tasks for different sample spaces, and to assess a model's general human-ness. 

In general, the human detection problem is an arms race between models and human detectors. Whenever a model becomes $ \tau $-undetectable for a given context, there will be a search for a transformation or human-conditional distribution or sample space for which the model is only $ \tau' $-undetectable for $ \tau' $ > $ \tau $. Along these lines, it may be useful to require models to produce ``watermarks" for the content they produce so as to make the detection problem much simpler \citep{kirchenbauer2023watermark}.

Finally, while the framework is presented in the context of human detection, it is possible to apply it to model-to-model comparisons in black box settings where only model input and model output are observed. The black-box setting is extremely limiting in terms of feasible types of model comparisons but is along a similar direction as previous work on similarity measures for partition based classification rules \citep{helm2020partitionbased}. Of immediate interest are investigations related to uncovering areas of the sample space in which foundation models produce meaningfully different outputs than corresponding quantized, distilled, or fine-tuned models.

\subsection*{Acknowledgements}
We would like to thank Greg Bowyer, Felicia K{\"o}rner, Spencer Rarrick, 
and Karl Tayeb
for their valuable suggestions, comments, and critiques.

\clearpage

\bibliographystyle{iclr2021/iclr2021_conference}
\bibliography{biblio}

\begin{thebibliography}{46}
\providecommand{\natexlab}[1]{#1}
\providecommand{\url}[1]{\texttt{#1}}
\expandafter\ifx\csname urlstyle\endcsname\relax
  \providecommand{\doi}[1]{doi: #1}\else
  \providecommand{\doi}{doi: \begingroup \urlstyle{rm}\Url}\fi

\bibitem[{Aaditya Bhat}(2023)]{aaditya_bhat_2023}
{Aaditya Bhat}.
\newblock Gpt-wiki-intro (revision 0e458f5), 2023.
\newblock URL
  \url{https://huggingface.co/datasets/aadityaubhat/GPT-wiki-intro}.

\bibitem[Anthropic(2023)]{Claude}
Anthropic.
\newblock Introducing claude, 2023.
\newblock URL \url{https://www.anthropic.com/index/introducing-claude}.

\bibitem[Bakhtin et~al.(2019)Bakhtin, Gross, Ott, Deng, Ranzato, and
  Szlam]{bakhtin2019real}
Anton Bakhtin, Sam Gross, Myle Ott, Yuntian Deng, Marc'Aurelio Ranzato, and
  Arthur Szlam.
\newblock Real or fake? learning to discriminate machine from human generated
  text, 2019.

\bibitem[Beaudry \& Renner(2012)Beaudry and Renner]{beaudry2012intuitive}
Normand~J. Beaudry and Renato Renner.
\newblock An intuitive proof of the data processing inequality, 2012.

\bibitem[Bickel \& Doksum(2015)Bickel and Doksum]{bickel2015mathematical}
Peter~J Bickel and Kjell~A Doksum.
\newblock \emph{Mathematical statistics: basic ideas and selected topics,
  volumes I-II package}.
\newblock CRC Press, 2015.

\bibitem[Bubeck et~al.(2023)Bubeck, Chandrasekaran, Eldan, Gehrke, Horvitz,
  Kamar, Lee, Lee, Li, Lundberg, Nori, Palangi, Ribeiro, and
  Zhang]{bubeck2023sparks}
Sébastien Bubeck, Varun Chandrasekaran, Ronen Eldan, Johannes Gehrke, Eric
  Horvitz, Ece Kamar, Peter Lee, Yin~Tat Lee, Yuanzhi Li, Scott Lundberg,
  Harsha Nori, Hamid Palangi, Marco~Tulio Ribeiro, and Yi~Zhang.
\newblock Sparks of artificial general intelligence: Early experiments with
  gpt-4, 2023.

\bibitem[Chen et~al.(2021)Chen, Tworek, Jun, Yuan, de~Oliveira~Pinto, Kaplan,
  Edwards, Burda, Joseph, Brockman, Ray, Puri, Krueger, Petrov, Khlaaf, Sastry,
  Mishkin, Chan, Gray, Ryder, Pavlov, Power, Kaiser, Bavarian, Winter, Tillet,
  Such, Cummings, Plappert, Chantzis, Barnes, Herbert-Voss, Guss, Nichol,
  Paino, Tezak, Tang, Babuschkin, Balaji, Jain, Saunders, Hesse, Carr, Leike,
  Achiam, Misra, Morikawa, Radford, Knight, Brundage, Murati, Mayer, Welinder,
  McGrew, Amodei, McCandlish, Sutskever, and Zaremba]{chen2021evaluating}
Mark Chen, Jerry Tworek, Heewoo Jun, Qiming Yuan, Henrique~Ponde
  de~Oliveira~Pinto, Jared Kaplan, Harri Edwards, Yuri Burda, Nicholas Joseph,
  Greg Brockman, Alex Ray, Raul Puri, Gretchen Krueger, Michael Petrov, Heidy
  Khlaaf, Girish Sastry, Pamela Mishkin, Brooke Chan, Scott Gray, Nick Ryder,
  Mikhail Pavlov, Alethea Power, Lukasz Kaiser, Mohammad Bavarian, Clemens
  Winter, Philippe Tillet, Felipe~Petroski Such, Dave Cummings, Matthias
  Plappert, Fotios Chantzis, Elizabeth Barnes, Ariel Herbert-Voss,
  William~Hebgen Guss, Alex Nichol, Alex Paino, Nikolas Tezak, Jie Tang, Igor
  Babuschkin, Suchir Balaji, Shantanu Jain, William Saunders, Christopher
  Hesse, Andrew~N. Carr, Jan Leike, Josh Achiam, Vedant Misra, Evan Morikawa,
  Alec Radford, Matthew Knight, Miles Brundage, Mira Murati, Katie Mayer, Peter
  Welinder, Bob McGrew, Dario Amodei, Sam McCandlish, Ilya Sutskever, and
  Wojciech Zaremba.
\newblock Evaluating large language models trained on code, 2021.

\bibitem[Chollet(2019)]{chollet2019measure}
Fran{\c{c}}ois Chollet.
\newblock On the measure of intelligence.
\newblock \emph{arXiv preprint arXiv:1911.01547}, 2019.

\bibitem[Devroye et~al.(2013)Devroye, Gy{\"o}rfi, and
  Lugosi]{devroye2013probabilistic}
Luc Devroye, L{\'a}szl{\'o} Gy{\"o}rfi, and G{\'a}bor Lugosi.
\newblock \emph{A probabilistic theory of pattern recognition}, volume~31.
\newblock Springer Science \& Business Media, 2013.

\bibitem[Duderstadt et~al.(2023)Duderstadt, Helm, and
  Priebe]{duderstadt2023comparing}
Brandon Duderstadt, Hayden~S. Helm, and Carey~E. Priebe.
\newblock Comparing foundation models using data kernels, 2023.

\bibitem[Eloundou et~al.(2023)Eloundou, Manning, Mishkin, and
  Rock]{eloundou2023gpts}
Tyna Eloundou, Sam Manning, Pamela Mishkin, and Daniel Rock.
\newblock Gpts are gpts: An early look at the labor market impact potential of
  large language models, 2023.

\bibitem[Fan et~al.(2018)Fan, Lewis, and Dauphin]{fan-etal-2018-hierarchical}
Angela Fan, Mike Lewis, and Yann Dauphin.
\newblock Hierarchical neural story generation.
\newblock In \emph{Proceedings of the 56th Annual Meeting of the Association
  for Computational Linguistics (Volume 1: Long Papers)}, pp.\  889--898,
  Melbourne, Australia, July 2018. Association for Computational Linguistics.
\newblock \doi{10.18653/v1/P18-1082}.
\newblock URL \url{https://aclanthology.org/P18-1082}.

\bibitem[Gehrmann et~al.(2019)Gehrmann, Strobelt, and
  Rush]{gehrmann-etal-2019-gltr}
Sebastian Gehrmann, Hendrik Strobelt, and Alexander Rush.
\newblock {GLTR}: Statistical detection and visualization of generated text.
\newblock In \emph{Proceedings of the 57th Annual Meeting of the Association
  for Computational Linguistics: System Demonstrations}, pp.\  111--116,
  Florence, Italy, July 2019. Association for Computational Linguistics.
\newblock \doi{10.18653/v1/P19-3019}.
\newblock URL \url{https://aclanthology.org/P19-3019}.

\bibitem[Geman et~al.(2015)Geman, Geman, Hallonquist, and
  Younes]{doi:10.1073/pnas.1422953112}
Donald Geman, Stuart Geman, Neil Hallonquist, and Laurent Younes.
\newblock Visual turing test for computer vision systems.
\newblock \emph{Proceedings of the National Academy of Sciences}, 112\penalty0
  (12):\penalty0 3618--3623, 2015.
\newblock \doi{10.1073/pnas.1422953112}.
\newblock URL \url{https://www.pnas.org/doi/abs/10.1073/pnas.1422953112}.

\bibitem[Google(2023)]{PaLM2}
Google.
\newblock Palm2 technical report, 2023.
\newblock URL \url{https://ai.google/static/documents/palm2techreport.pdf}.

\bibitem[GPT4-AOAI()]{GPT4-AOAI}
GPT4-AOAI.
\newblock Azure openai gpt-4 (preview).
\newblock
  \url{https://azure.microsoft.com/en-us/products/cognitive-services/openai-service},
  2023.

\bibitem[Helm et~al.(2020)Helm, Mehta, Duderstadt, Yang, White, Geisa,
  Vogelstein, and Priebe]{helm2020partitionbased}
Hayden~S. Helm, Ronak~D. Mehta, Brandon Duderstadt, Weiwei Yang, Christoper~M.
  White, Ali Geisa, Joshua~T. Vogelstein, and Carey~E. Priebe.
\newblock A partition-based similarity for classification distributions, 2020.
\newblock URL \url{https://arxiv.org/abs/2011.06557}.

\bibitem[Hern{\'a}ndez-Orallo(2017)]{hernandez2017measure}
Jos{\'e} Hern{\'a}ndez-Orallo.
\newblock \emph{The measure of all minds: evaluating natural and artificial
  intelligence}.
\newblock Cambridge University Press, 2017.

\bibitem[Huang(2023)]{nytimesUniversityAIBan}
Kalley Huang.
\newblock Alarmed by a.i. chatbots, universities start revamping how they teach
  with the rise of the popular new chatbot chatgpt, colleges are restructuring
  some courses and taking preventive measures.
\newblock
  \url{https://www.nytimes.com/2023/01/16/technology/chatgpt-artificial-intelligence-universities.html},
  2023.
\newblock [Accessed 15-May-2023].

\bibitem[Ippolito et~al.(2020)Ippolito, Duckworth, Callison-Burch, and
  Eck]{ippolito-etal-2020-automatic}
Daphne Ippolito, Daniel Duckworth, Chris Callison-Burch, and Douglas Eck.
\newblock Automatic detection of generated text is easiest when humans are
  fooled.
\newblock In \emph{Proceedings of the 58th Annual Meeting of the Association
  for Computational Linguistics}, pp.\  1808--1822, Online, July 2020.
  Association for Computational Linguistics.
\newblock \doi{10.18653/v1/2020.acl-main.164}.
\newblock URL \url{https://aclanthology.org/2020.acl-main.164}.

\bibitem[Jin et~al.(2019)Jin, Dhingra, Liu, Cohen, and
  Lu]{jin-etal-2019-pubmedqa}
Qiao Jin, Bhuwan Dhingra, Zhengping Liu, William Cohen, and Xinghua Lu.
\newblock {P}ub{M}ed{QA}: A dataset for biomedical research question answering.
\newblock In \emph{Proceedings of the 2019 Conference on Empirical Methods in
  Natural Language Processing and the 9th International Joint Conference on
  Natural Language Processing (EMNLP-IJCNLP)}, pp.\  2567--2577, Hong Kong,
  China, November 2019. Association for Computational Linguistics.
\newblock \doi{10.18653/v1/D19-1259}.
\newblock URL \url{https://aclanthology.org/D19-1259}.

\bibitem[Khanjani et~al.(2023)Khanjani, Watson, and Janeja]{khanjani2023audio}
Zahra Khanjani, Gabrielle Watson, and Vandana~P Janeja.
\newblock Audio deepfakes: A survey.
\newblock \emph{Frontiers in Big Data}, 5:\penalty0 1001063, 2023.

\bibitem[Kirchenbauer et~al.(2023)Kirchenbauer, Geiping, Wen, Katz, Miers, and
  Goldstein]{kirchenbauer2023watermark}
John Kirchenbauer, Jonas Geiping, Yuxin Wen, Jonathan Katz, Ian Miers, and Tom
  Goldstein.
\newblock A watermark for large language models, 2023.

\bibitem[Liang et~al.(2022)Liang, Bommasani, Lee, Tsipras, Soylu, Yasunaga,
  Zhang, Narayanan, Wu, Kumar, Newman, Yuan, Yan, Zhang, Cosgrove, Manning,
  Ré, Acosta-Navas, Hudson, Zelikman, Durmus, Ladhak, Rong, Ren, Yao, Wang,
  Santhanam, Orr, Zheng, Yuksekgonul, Suzgun, Kim, Guha, Chatterji, Khattab,
  Henderson, Huang, Chi, Xie, Santurkar, Ganguli, Hashimoto, Icard, Zhang,
  Chaudhary, Wang, Li, Mai, Zhang, and Koreeda]{helm_not_hayden}
Percy Liang, Rishi Bommasani, Tony Lee, Dimitris Tsipras, Dilara Soylu,
  Michihiro Yasunaga, Yian Zhang, Deepak Narayanan, Yuhuai Wu, Ananya Kumar,
  Benjamin Newman, Binhang Yuan, Bobby Yan, Ce~Zhang, Christian Cosgrove,
  Christopher~D. Manning, Christopher Ré, Diana Acosta-Navas, Drew~A. Hudson,
  Eric Zelikman, Esin Durmus, Faisal Ladhak, Frieda Rong, Hongyu Ren, Huaxiu
  Yao, Jue Wang, Keshav Santhanam, Laurel Orr, Lucia Zheng, Mert Yuksekgonul,
  Mirac Suzgun, Nathan Kim, Neel Guha, Niladri Chatterji, Omar Khattab, Peter
  Henderson, Qian Huang, Ryan Chi, Sang~Michael Xie, Shibani Santurkar, Surya
  Ganguli, Tatsunori Hashimoto, Thomas Icard, Tianyi Zhang, Vishrav Chaudhary,
  William Wang, Xuechen Li, Yifan Mai, Yuhui Zhang, and Yuta Koreeda.
\newblock Holistic evaluation of language models, 2022.
\newblock URL \url{https://arxiv.org/abs/2211.09110}.

\bibitem[Liu et~al.(2021)Liu, Yuan, Fu, Jiang, Hayashi, and
  Neubig]{liu2021pretrain}
Pengfei Liu, Weizhe Yuan, Jinlan Fu, Zhengbao Jiang, Hiroaki Hayashi, and
  Graham Neubig.
\newblock Pre-train, prompt, and predict: A systematic survey of prompting
  methods in natural language processing, 2021.

\bibitem[Longpre et~al.(2023)Longpre, Yauney, Reif, Lee, Roberts, Zoph, Zhou,
  Wei, Robinson, Mimno, and Ippolito]{longpre2023pretrainers}
Shayne Longpre, Gregory Yauney, Emily Reif, Katherine Lee, Adam Roberts, Barret
  Zoph, Denny Zhou, Jason Wei, Kevin Robinson, David Mimno, and Daphne
  Ippolito.
\newblock A pretrainer's guide to training data: Measuring the effects of data
  age, domain coverage, quality, \& toxicity, 2023.

\bibitem[Mitchell et~al.(2023)Mitchell, Lee, Khazatsky, Manning, and
  Finn]{mitchell2023detectgpt}
Eric Mitchell, Yoonho Lee, Alexander Khazatsky, Christopher~D. Manning, and
  Chelsea Finn.
\newblock Detectgpt: Zero-shot machine-generated text detection using
  probability curvature, 2023.
\newblock URL \url{https://arxiv.org/abs/2301.11305}.

\bibitem[Narayan et~al.(2018)Narayan, Cohen, and
  Lapata]{narayan-etal-2018-dont}
Shashi Narayan, Shay~B. Cohen, and Mirella Lapata.
\newblock Don{'}t give me the details, just the summary! topic-aware
  convolutional neural networks for extreme summarization.
\newblock In \emph{Proceedings of the 2018 Conference on Empirical Methods in
  Natural Language Processing}, pp.\  1797--1807, Brussels, Belgium,
  October-November 2018. Association for Computational Linguistics.
\newblock \doi{10.18653/v1/D18-1206}.
\newblock URL \url{https://aclanthology.org/D18-1206}.

\bibitem[Nguyen et~al.(2022)Nguyen, Nguyen, Nguyen, Nguyen, Huynh-The,
  Nahavandi, Nguyen, Pham, and Nguyen]{nguyen2022deep}
Thanh~Thi Nguyen, Quoc Viet~Hung Nguyen, Dung~Tien Nguyen, Duc~Thanh Nguyen,
  Thien Huynh-The, Saeid Nahavandi, Thanh~Tam Nguyen, Quoc-Viet Pham, and
  Cuong~M Nguyen.
\newblock Deep learning for deepfakes creation and detection: A survey.
\newblock \emph{Computer Vision and Image Understanding}, 223:\penalty0 103525,
  2022.

\bibitem[OpenAI(2023{\natexlab{a}})]{AITextClassifier}
OpenAI, January 2023{\natexlab{a}}.
\newblock URL \url{https://beta.openai.com/ai-text-classifier}.

\bibitem[OpenAI(2023{\natexlab{b}})]{gpt4systemcard}
OpenAI.
\newblock Gpt-4 system card, 2023{\natexlab{b}}.
\newblock URL \url{https://cdn.openai.com/papers/gpt-4-system-card.pdf}.

\bibitem[Powers(1998)]{powers1998total}
David~MW Powers.
\newblock The total turing test and the loebner prize.
\newblock In \emph{New Methods in Language Processing and Computational Natural
  Language Learning}, 1998.

\bibitem[Radford et~al.(2019)Radford, Wu, Child, Luan, Amodei, and
  Sutskever]{radford2019language}
Alec Radford, Jeff Wu, Rewon Child, David Luan, Dario Amodei, and Ilya
  Sutskever.
\newblock Language models are unsupervised multitask learners.
\newblock 2019.

\bibitem[Radford et~al.(2021)Radford, Kim, Hallacy, Ramesh, Goh, Agarwal,
  Sastry, Askell, Mishkin, Clark, Krueger, and Sutskever]{clip}
Alec Radford, Jong~Wook Kim, Chris Hallacy, Aditya Ramesh, Gabriel Goh,
  Sandhini Agarwal, Girish Sastry, Amanda Askell, Pamela Mishkin, Jack Clark,
  Gretchen Krueger, and Ilya Sutskever.
\newblock Learning transferable visual models from natural language
  supervision.
\newblock \emph{CoRR}, abs/2103.00020, 2021.
\newblock URL \url{https://arxiv.org/abs/2103.00020}.

\bibitem[Radford et~al.(2022)Radford, Kim, Xu, Brockman, McLeavey, and
  Sutskever]{radford2022robust}
Alec Radford, Jong~Wook Kim, Tao Xu, Greg Brockman, Christine McLeavey, and
  Ilya Sutskever.
\newblock Robust speech recognition via large-scale weak supervision, 2022.

\bibitem[Raffel et~al.(2020)Raffel, Shazeer, Roberts, Lee, Narang, Matena,
  Zhou, Li, and Liu]{raffel2020exploring}
Colin Raffel, Noam Shazeer, Adam Roberts, Katherine Lee, Sharan Narang, Michael
  Matena, Yanqi Zhou, Wei Li, and Peter~J. Liu.
\newblock Exploring the limits of transfer learning with a unified text-to-text
  transformer, 2020.

\bibitem[Satariano(2023)]{nytimesChatGPTBanned}
Adam Satariano.
\newblock {C}hat{G}{P}{T} {I}s {B}anned in {I}taly {O}ver {P}rivacy {C}oncerns
  --- nytimes.com.
\newblock
  \url{https://www.nytimes.com/2023/03/31/technology/chatgpt-italy-ban.html},
  2023.
\newblock [Accessed 14-May-2023].

\bibitem[Saveri et~al.(2023)Saveri, Zirpoli, Young, Buchanan, and
  Manfredi]{stable-diffusion-litigation}
Joseph~R. Saveri, Cadio Zirpoli, Christopher~K.L. Young, Elissa~A. Buchanan,
  and Travis Manfredi.
\newblock class-action lawsuit against midourney and stable ai, 2023.
\newblock URL
  \url{https://stablediffusionlitigation.com/pdf/00201/1-1-stable-diffusion-complaint.pdf}.

\bibitem[Solaiman et~al.(2019)Solaiman, Brundage, Clark, Askell, Herbert-Voss,
  Wu, Radford, Krueger, Kim, Kreps, McCain, Newhouse, Blazakis, McGuffie, and
  Wang]{solaiman2019release}
Irene Solaiman, Miles Brundage, Jack Clark, Amanda Askell, Ariel Herbert-Voss,
  Jeff Wu, Alec Radford, Gretchen Krueger, Jong~Wook Kim, Sarah Kreps, Miles
  McCain, Alex Newhouse, Jason Blazakis, Kris McGuffie, and Jasmine Wang.
\newblock Release strategies and the social impacts of language models, 2019.

\bibitem[Song et~al.(2020)Song, Tan, Qin, Lu, and Liu]{song2020mpnet}
Kaitao Song, Xu~Tan, Tao Qin, Jianfeng Lu, and Tie-Yan Liu.
\newblock Mpnet: Masked and permuted pre-training for language understanding,
  2020.

\bibitem[TURING(1950)]{10.1093/mind/LIX.236.433}
A.~M. TURING.
\newblock {I.—COMPUTING MACHINERY AND INTELLIGENCE}.
\newblock \emph{Mind}, LIX\penalty0 (236):\penalty0 433--460, 10 1950.
\newblock ISSN 0026-4423.
\newblock \doi{10.1093/mind/LIX.236.433}.
\newblock URL \url{https://doi.org/10.1093/mind/LIX.236.433}.

\bibitem[Uchendu et~al.(2020)Uchendu, Le, Shu, and
  Lee]{uchendu-etal-2020-authorship}
Adaku Uchendu, Thai Le, Kai Shu, and Dongwon Lee.
\newblock Authorship attribution for neural text generation.
\newblock In \emph{Proceedings of the 2020 Conference on Empirical Methods in
  Natural Language Processing (EMNLP)}, pp.\  8384--8395, Online, November
  2020. Association for Computational Linguistics.
\newblock \doi{10.18653/v1/2020.emnlp-main.673}.
\newblock URL \url{https://aclanthology.org/2020.emnlp-main.673}.

\bibitem[Verdoliva(2020)]{Verdoliva_2020}
Luisa Verdoliva.
\newblock Media forensics and {DeepFakes}: An overview.
\newblock \emph{{IEEE} Journal of Selected Topics in Signal Processing},
  14\penalty0 (5):\penalty0 910--932, aug 2020.
\newblock \doi{10.1109/jstsp.2020.3002101}.
\newblock URL \url{https://doi.org/10.1109%2Fjstsp.2020.3002101}.

\bibitem[Weiser(2023)]{lawyers-chat-gpt}
Benjamin Weiser.
\newblock Chatgpt lawyers are ordered to consider seeking forgiveness.
\newblock \emph{The New York Times}, 2023.
\newblock URL
  \url{https://www.nytimes.com/2023/06/22/nyregion/lawyers-chatgpt-schwartz-loduca.html,
  urldate = {2023-08-20},}.

\bibitem[Wolf et~al.(2019)Wolf, Debut, Sanh, Chaumond, Delangue, Moi, Cistac,
  Rault, Louf, Funtowicz, et~al.]{wolf2019huggingface}
Thomas Wolf, Lysandre Debut, Victor Sanh, Julien Chaumond, Clement Delangue,
  Anthony Moi, Pierric Cistac, Tim Rault, R{\'e}mi Louf, Morgan Funtowicz,
  et~al.
\newblock Huggingface's transformers: State-of-the-art natural language
  processing.
\newblock \emph{arXiv preprint arXiv:1910.03771}, 2019.

\bibitem[Zhang et~al.(2022)Zhang, Dellaferrera, Sikarwar, Armendariz, Mudrik,
  Agrawal, Madan, Barbu, Yang, Kumar, Sadwani, Dellaferrera, Pizzochero,
  Pfister, and Kreiman]{zhang2022human}
Mengmi Zhang, Giorgia Dellaferrera, Ankur Sikarwar, Marcelo Armendariz, Noga
  Mudrik, Prachi Agrawal, Spandan Madan, Andrei Barbu, Haochen Yang, Tanishq
  Kumar, Meghna Sadwani, Stella Dellaferrera, Michele Pizzochero, Hanspeter
  Pfister, and Gabriel Kreiman.
\newblock Human or machine? turing tests for vision and language, 2022.

\end{thebibliography}

\clearpage

\appendix

\section*{Algorithmic description of ProxiHuman}

\begin{algorithm}[h!]
\caption{ProxiHuman}\label{alg:Proxihuman}
\begin{algorithmic}
\Require 
Content $ x $, perturbation function $ \tilde{p} $, embedding function $ g $, number of pertubrations $ n_{q} $, distance function $ d $, number of neighbors $ k $, threshold $ c $.
\State $ y \gets g(x) $ \Comment{Embed $ x $}
\For{$ i \in \{1, 2, \hdots, n_{q}\} $}
\State $ \tilde{x}_{i} \sim \tilde{p}(x) $ \Comment{Sample from $ \tilde{p}(x) $}
\State $ y_{i} \gets g(\tilde{x}_{i})$ \Comment{Embed $ \tilde{x}_{i} $}
\EndFor
\State $ \texttt{dist} \gets \frac{1}{k} \sum_{i=1}^{n_{q}} d(y, y_{i}) \cdot \mathbbm{1}\{y_{i} \in \mathcal{N}_{k}(y, d)\} $ \Comment{Store average distance to $ k $-NN}
\If{$ \texttt{dist} > c$} 
\State \Return \texttt{True} \Comment{The average distance is above a threshold, return Human}
\EndIf
\State \Return \texttt{False} 
\end{algorithmic}
\end{algorithm}

\end{document}